# Trimming Phonetic Alignments Improves the Inference of Sound Correspondence Patterns from Multilingual Wordlists

**Frederic Blum**
DLCE
MPI-EVA
Leipzig, Germany
`frederic_blum@eva.mpg.de`

**Johann-Mattis List**
Chair of Multil. Comput. Linguistics / DLCE
University of Passau / MPI-EVA
Passau / Leipzig, Germany
`mattis.list@uni-passau.de`

## Abstract

Sound correspondence patterns form the basis of cognate detection and phonological reconstruction in historical language comparison. Methods for the automatic inference of correspondence patterns from phonetically aligned cognate sets have been proposed, but their application to multilingual wordlists requires extremely well annotated datasets. Since annotation is tedious and time consuming, it would be desirable to find ways to improve aligned cognate data automatically. Taking inspiration from trimming techniques in evolutionary biology, which improve alignments by excluding problematic sites, we propose a workflow that trims phonetic alignments in comparative linguistics prior to the inference of correspondence patterns. Testing these techniques on a large standardized collection of ten datasets with expert annotations from different language families, we find that the best trimming technique substantially improves the overall consistency of the alignments. The results show a clear increase in the proportion of frequent correspondence patterns and words exhibiting regular cognate relations.

## 1 Introduction

With the introduction of automated methods for the inference of correspondence patterns from multilingual wordlists (List, 2019), computational historical linguistics has acquired a new technique with multiple applications in the field. Correspondence patterns have been used to identify problematic cognate judgments in individual datasets (List, 2019) or to assess their general characteristics (Wu et al., 2020), they have been used as the basis to predict cognate reflexes (Bodt and List, 2022; List et al., 2022c; Tresoldi et al., 2022) or to reconstruct proto-forms (List et al., 2022b). They have also shown to be useful to compare different cognate judgments with respect to the overall regularity they introduce in a multilingual dataset (Greenhill et al., 2023).

While machine-readable correspondence patterns have already shown to be useful for various tasks in historical linguistics, their basic properties have so far not yet been thoroughly investigated. Thus, although we can easily see that correspondence patterns show long-tail distributions with respect to the number of alignment sites that individual patterns reflect in multilingual datasets, no closer investigations of these patterns have been carried out so far. Here, historical linguistics can learn from evolutionary biology, where specific characteristics of alignments of DNA or protein sequences have been investigated for several decades now. Scholars have also looked into the characteristics of those alignment sites that turn out to be problematic when it comes to phylogenetic reconstruction and similar secondary tasks (Talavera and Castresana, 2007; Dress et al., 2008). In order to handle these "irregular" sites, biologists have proposed methods to *trim* alignments by removing sites that contradict more general evolutionary tendencies. This allows scholars to reduce the amount of artifacts in the data and retrieve more accurate information about the evolutionary processes behind the alignments.

In computational historical linguistics, *trimming* of alignments has so far been ignored. In classical historical language comparison, however, the practice of ignoring specific sites in the alignment of cognate words has been practiced for a long time. When arguing for particular sound changes or correspondence patterns, scholars routinely consider only the supposed *root* of a cognate set (Trask, 2000, 290), ignoring inflectional and derivational markers or irregular parts of individual cognate reflexes. While this is a common procedure for the comparative method, it is seldom made explicit. One of the few cases where this process *is* made explicit is offered by Payne (1991). Here, the author provides an alignment matrix where all the non-cognate material is set into brackets, distin-guishing them from the true alignment sites. This step is accompanied by a detailed discussion of the morphemic elements and its implication for reconstructing the proto-forms, a step that is rarely put into such detail. The importance of this practice is also reflected in tools that allow for the manual correction of alignments, like EDICTOR (List, 2017a) and RefLex (Segerer and Flavier, 2015) which offer options to flag alignment sites as problematic (or important). Specifically the trimming facility of the EDICTOR tool has also been used to increase the transparency of cognate sets in studies devoted to phylogenetic reconstruction (Sagart et al., 2019; Cayón and Chacon, 2022).

Given the highly skewed distributions of alignment sites over correspondence patterns in computational comparative linguistics and the practice of human annotators to regularly ignore certain parts of phonetically aligned cognate sets in historical linguistics, it would be beneficial to find automated ways to *trim* phonetic alignments in multilingual wordlists. Trimmed alignments could either form the basis of a more extensive annotation of phonetic alignments in a computer-assisted setting (List, 2017b), or they could serve as the basis of extensive cross-linguistic, typologically oriented studies devoted to the regularity of sound change and sound correspondence patterns. For example, correspondence patterns have already been used in typological studies investigating the history of pronoun systems in South America (Rojas-Berscia and Roberts, 2020), or for studies with simulated data that use phonetic alignments to construct artificial cognate sets (Wichmann and Rama, 2021).

In the following, we will provide a first framework for the trimming of phonetic alignments and test it on ten datasets from typologically diverse language families. Our experiments show that trimming increases the overall regularity of the correspondence patterns – even when using very rudimentary strategies– and thus shrinks the long tail of their distributions over alignment sites. The closer inspection of individual trimmed alignments, however, also shows that our methods still have a lot of room for improvement. We conclude by pointing to various techniques that could enhance the trimming of phonetic alignments in the future.

## 2 Background

Sound correspondences are the core of the comparative method. They form the basis for proving genetic relationship between languages, for establishing the internal classification of language families, as well as for the reconstruction of proto-languages. Sets of sound correspondences are commonly analyzed as *correspondence patterns*. A crucial component of correspondence patterns in contrast to sound correspondences is that the correspondence set is not defined on the basis of language pairs, but rather as a pattern shared between several languages (List, 2019, 141). In other words, a correspondence pattern is defined as the set of sounds in any number of daughter languages that derive from the same phoneme of the ancestral language in a specific environment (Hoenigswald, 1960; Anttila, 1972).

In order to qualify as a *pattern*, sound correspondences must be backed by many examples. Examples are drawn from concrete cognate sets that need to be phonetically aligned in order to reveal which sounds correspond with each other. In order to constitute a valid pattern that would be accepted as a *regular* or *systematic* sound correspondence (Trask, 2000, 336f), a considerably large amount of examples backing a certain pattern must be assembled from the data. This step is necessary to avoid chance similarities resulting from erroneous cognate judgments or undetected scarce borrowings. While the minimum number of examples is not universally agreed upon, most scholars tend to accept two or three examples as sufficient to consider a pattern as regular.

Correspondence patterns are typically represented with the help of a matrix, in which the rows correspond to individual languages and the columns correspond to patterns, with cell values indicating the sounds (the *reflexes*) of individual language varieties in individual patterns (Clackson, 2007, 307). Correspondence patterns are traditionally inferred by manually inspecting phonetic alignments of cognate sets, trying to identify individual columns (*alignment sites*) in the alignments that are compatible with each other (Anttila, 1972; List, 2019). Figure 1 illustrates this process with phonetic alignments of fictitious words from fictitious languages. In order to reconstruct the ancestral form underlying a cognate set, it is common to ignore certain sites in the alignment that are considered as difficult to align. Problems of alignability (Schweikhard and List, 2020, 10) usually result from the fact that words in a cognate set are not entirely, but only partially cognate. This can be

|            | I  |   | II |   | II |   | I  |   |
|------------|----|---|----|---|----|---|----|---|
| Language A | t  | a | h  | e | h  | i | t  | u |
| Language B | tʰ | a | x  | e | x  | u | tʰ | i |
| Language C | t  | a | x  | e | x  | u | t  | i |
| Language D | ts | a | x  | e | x  | u | ts | i |

Figure 1: Corresponding alignment sites in a set of four fictitious languages.

| Pacaraos | w | a | ɲ | u | + | k | u |
| Napo     | w | a | ɲ | u | + | n | a |
| Pastaza  | w | a | ɲ | u | + | n | a |
| Ayacucho | w | a | ɲ | u |   |   |   |
| Jauja    | w | a | ɲ | u |   |   |   |
| Lamas    | w | a | ɲ | u |   |   |   |

Figure 2: Trimming morphemes in Quechua. The root is combined with different morphemes in some varieties.

due to processes of word formation or inflection in individual language varieties (Wu and List, 2023), as illustrated in Figure 2 with data from Quechua (Blum et al., forthcoming).

## 3 Materials and Methods

### 3.1 Materials

We use ten freely available datasets from typologically diverse language families, taken from the Lexibank collection (List et al., 2022a). This collection contains datasets that were (retro)standardized following the recommendations of the Cross-Linguistic Data Formats initiative (CLDF, https://cldf.clld.org, Forkel et al. 2018). One core aspect of CLDF is to make active use of *reference catalogs* like Glottolog (https://glottolog.org, Hammarström et al. 2022) and Concepticon (https://concepticon.clld.org), List et al. 2023). Reference catalogs in this context are metadata collections that provide extensive information on very general linguistic constructs, such as languages, concepts, or speech sounds. By linking the languages in a given dataset to Glottolog, by providing Glottocodes for individiual language varieties, one guarantees the comparability of the language varieties with other datasets which have also been linked to Glottolog. By mapping concepts in multilingual wordlists to Concepticon, one guarantees the comparability of the concepts with other datasets that have also been linked to Concepticon. Apart from Glottolog and Concepticon, many datasets from the Lexibank collection offer standardized phonetic transcriptions following the Cross-Linguistic Transcription Systems reference catalog (CLTS, https://clts.clld.org, List et al. 2021, see Anderson et al. 2018). In this reference catalog, more than 8000 different speech sounds are defined and can be distinguished with the help of distinctive features. At the same time, new, so far unseen sounds can be derived using a specific parsing algorithm underlying the PyCLTS software package (List et al., 2020). As a result, the Lexibank collection of multilingual wordlists offers a large number of multilingual datasets that have been standardized with respect to languages, concepts, and transcriptions.

Apart from offering standardized phonetic transcriptions, all datasets also offer cognate judgments provided by experts. Alignments were computed automatically, using the SCA method for multiple phonetic alignments (List, 2012, 2014) in its default settings. Of the ten datasets, two (CROSSANDEAN and WALWORTHPOLYNESIAN) were reduced to 20 language varieties in order to have datasets of comparable sizes. While the datasets differ with respect to the number of language varieties and time depth of the families in question, they are all large enough to allow us to infer a substantial amount of frequent sound correspondence patterns.

### 3.2 Methods

#### 3.2.1 Trimming Phonetic Alignments

The main purpose of trimming is to remove problematic alignments and increase the potential of retrieving relevant information from the remaining sites. In biology, trimming of sequence alignments is primarily performed to improve phylogenetic inference. The goal is to reduce the noise in the data in order to get a clearer picture of the actual phylogenetic information contained in DNA sequences (Talavera and Castresana, 2007). Despite the removal of some data, the accuracy of phylogenetic trees inferred from the data often improves. To assure that enough relevant information is maintained after trimming, trimmed alignments need to have some minimal length. Several tools for automated trimming have been developed in evolutionary biology. Some of them select the most reliable columns and remove sparse alignments that consists mainly of gaps (Capella-Gutiérrez et al., 2009), while other tools focus on entropy values and evaluate whether a site is expected or not (Criscuolo and Gribaldo, 2010). The most ambiguous and divergent sites are removed in this approach, arguing that they might result from erroneous judgements of homology (Steenwyk et al., 2020).

| Data set | Lang. | Concepts | Cog.-Sets | Words | Source |
|---|---|---|---|---|---|
| CONSTENLACHIBCHAN | 25 | 106 | 213 | 1216 | Constenla Umaña (2005) |
| CROSSANDEAN | 20 | 150 | 223 | 2789 | Blum et al. (forthcoming) |
| DRAVLEX | 20 | 100 | 179 | 1341 | Kolipakam et al. (2018) |
| FELEKESEMITIC | 21 | 150 | 271 | 2622 | Feleke (2021) |
| HATTORIJAPONIC | 10 | 197 | 235 | 1710 | Hattori (1973) |
| HOUCHINESE | 15 | 139 | 228 | 1816 | Hóu (2004) |
| LEEKOREANIC | 15 | 206 | 233 | 2131 | Lee (2015) |
| ROBINSONAP | 13 | 216 | 253 | 1424 | Robinson and Holton (2012) |
| WALWORTHPOLYNESIAN | 20 | 205 | 383 | 3637 | Walworth (2018) |
| ZHIVLOVOBUGRIAN | 21 | 110 | 182 | 1974 | Zhivlov (2011) |

Table 1: Number of languages, concepts, non-singleton cognate sets and total entries across the different datasets

In contrast to the trimming of DNA sequences in biology, the main goal of trimming alignments in linguistics is not to infer phylogenetic trees, but to make the alignments more useful for secondary use in computing sound correspondences and helping phonological reconstruction. Each cognate set is reduced to a 'core' alignment, which can then later be reconstructed as approximating the *root* in the proto-language of the respective cognate set.

Our initial trimming strategies focus on the presence of gaps in the alignment sites. For this purpose, we compute the proportion of gaps in each site and evaluate whether this proportion is above or below a certain threshold (*gap threshold*). All sites which are above the threshold are identified as *candidates* for trimming. The default value for the gap threshold in our implementation is 0.5, which means that we *could* trim all sites in which the majority of sounds is a gap.

However, since a naive trimming of all alignment sites exceeding our gap threshold might well lead to the trimming of all sites in an alignment and therefore discard the corresponding cognate set in its entirety, we define a minimal skeleton of alignment sites that should not be touched by the trimming procedure (similar to the minimal sequence length in DNA trimming). This skeleton is based on consonant-vowel profiles of the alignments and defaults to CV and VC. The preference of minimal CV/VC skeletons for aligned cognate sets is justified by linguistic practice (Tian et al., 2022) and can be adjusted to account for extended root structures, such as, for example, CVC . This means that only those results of the trimming procedure are accepted that leave a core alignment of at least one consonant and one vowel, ignoring their particular order. In order to make sure that the core is preserved, we first define an ordered list of candidate sites that could be removed and then start removing them site after site, checking after each removal whether the core skeleton has been left untouched. When only the core skeleton is left, trimming is stopped.

Based on this general procedure of trimming until a core skeleton defined by the user is reached, we test two detailed strategies for trimming. In the first strategy, we only trim *consecutive* gaps occurring in the beginning or the end of the alignment, a strategy that is also used in the context of sequence comparison in biology (Raghava and Barton, 2006). This *core-oriented* strategy allows us to drop spurious prefixes and suffixes occurring in some language varieties in individual alignments. In order to create our ordered list of candidate sites, we start from the right-most sites in our alignment and combine them with the left-most sites. In the second strategy, we trim all sites where the frequency of gaps exceeds our threshold, regardless of their position. This *gap-oriented* strategy would also trim gapped sites occurring in the beginning and the end of an alignment, but may additionally trim gapped sites regardless of their position. In order to create our ordered list of candidate sites, we sort all sites exceeding the gap threshold by the proportion of gaps in reversed order. Figure 3 illustrates the calculation of gap profiles and the trimming using the two strategies defined here for a toy example of fictitious words from fictitious languages.

| Language | Core-oriented | | | | | | | Gap-oriented | | | | | | |
|---|---|---|---|---|---|---|---|---|---|---|---|---|---|---|
| Language A | s | - | t | e | r | b | - | s | - | t | e | r | b | - |
| Language B | m | e | tʰ | e | - | - | - | m | e | tʰ | e | - | - | - |
| Language C | - | a | t | e | - | b | u | - | a | t | e | - | b | u |
| Language D | - | - | t | e | - | b | - | - | - | t | e | - | b | - |
| Gap proportion | 0.5 | 0.5 | 0.0 | 0.0 | 0.75 | 0.25 | 0.75 | 0.5 | 0.5 | 0.0 | 0.0 | 0.75 | 0.25 | 0.75 |

Figure 3: Artificial example for the computation of gap profiles followed by trimming using the *core-oriented* (left) and the *gap-oriented* strategy (right).

### 3.2.2 Evaluating Cognate Set Regularity

With the method by List (2019), correspondence patterns can be inferred from phonetically aligned cognate sets with the help of an iterative partitioning strategy which clusters the individual alignment sites. The resulting patterns are reflected by varying amounts of alignment sites, which we can use to compute certain statistics, building on earlier work by Greenhill et al. (2023). In a first step, we can compare the number of frequently recurring patterns with the number of patterns that do not recur frequently in the data. Based on this comparison, we can compute the proportion of alignment sites that are assigned to a frequently recurring pattern. This comes close to the notion of "regular" correspondence patterns in traditional historical linguistics, with the difference that we need to choose a concrete threshold by which a pattern recurs in our data (the *pattern threshold*, which is set to 3 by default). By defining frequently recurring patterns as *regular*, we can now assess for individual cognate sets how many of the alignment sites reflect regular patterns and how many reflect irregular patterns. This allows us to distinguish *regular* from *irregular* cognate sets by calculating the proportion of alignment sites reflecting regular correspondence patterns and setting some threshold beyond which we consider a cognate set as irregular (the *cognate threshold*, which is set to 0.75 by default). Having identified regular cognates in a given wordlist, we can contrast them with irregular cognates and calculate the proportion of *reflexes* (words in individual cognate sets) that appear in regular cognate sets. Given that this proportion gives us an idea of how many of the words in our data that appear in cognate relations can be assigned to some regular cognate set via regular sound correspondences, we interpret this proportion of *regular words* as the *overall regularity* of the dataset.

Selecting meaningful thresholds is not an easy task, specifically when calculations depend on multiple parameters as in our case. We decided to take a conservative pattern threshold of 3, which means that a pattern to be considered as regular must at least recur across three alignment sites in a given dataset. For the regularity of cognate sets, we decided for an even more conservative threshold of 0.75, which means that three quarters of the alignment sites in a given cognate set must reflect correspondence patterns that recur three or more times in the data.

### 3.2.3 Evaluating Trimmed Alignments

We make use of this interpretation of frequency as regularity in order to evaluate the success of our trimming operations. In order to check to which degree the trimming of phonetic alignments leads to an increase of overall regularity, modeled by taking the frequency of correspondence patterns into account, we compare three different constellations, namely (a) no trimming, (b) core-oriented trimming, and (c) gap-oriented trimming. We compare the three methods by computing the *proportion of regular correspondence patterns* and the *proportion of regular words* in all datasets, as outlined in the previous section. A successful trimming strategy should lead to an increase of both measures.

For further evaluation, we implement a random model that compares our targeted trimming strategies with a random strategy for trimming. To account for this, we randomly delete the same amount of alignment sites from each alignment as we did with the gap- or core-oriented strategies, while preserving the ratio of consonantal and vocalic alignment sites. With this step we assure that the resulting randomly trimmed alignment preserves the minimal CV/VC skeleton. For each dataset and trimming-strategy, we run the random model 100 times and analyze how many times the random model surpasses the results of the targeted model with respect to the proportion of regular words. This error analysis helps us to assess whether a trimming strategy systematically outperforms the random model.

### 3.2.4 Implementation

The new methods for the trimming of phonetic alignments are implemented in Python in the form of a plugin to the LingRex software package (https://pypi.org/project/lingrex, List and Forkel 2022, Version 1.3.0). LingRex itself extends LingPy (https://pypi.org/project/lingpy, List and Forkel 2021, Version 2.6.9) – which we use for phonetic alignments – by providing the method for correspondence pattern detection which we use to evaluate the consequences of trimming our alignments. For the handling of the cross-linguistic datasets provided in CLDF, CLDFBench (https://pypi.org/project/cldfbench, Forkel and List 2020, Version 1.13.0) is used with the PyLexibank plugin (https://pypi.org/project/pylexibank, Forkel et al. 2021, Version 3.4.0 ).

## 4 Results

### 4.1 General Results

The two trimming strategies were applied to all datasets in our sample and regularity scores for the proportion of regular sound correspondence patterns and the proportion of regular words were computed. Given that the trimming strategies might reduce alignments only to a core skeleton (CV/VC), only those cognate sets whose alignments consist of at least one vocalic and one consonantal site were considered in this comparison. Phonetic alignments were carried out with the help of the default settings of the SCA method (List, 2012). Correspondence patterns were computed with the help of the method by List (2019). The results of our general comparison of different trimming strategies are presented in Table 2. For both the proportion of regular correspondence patterns and the proportion of regular words, the best result for each dataset is highlighted in the table. Without exception, the gap-oriented trimming strategy yields the highest proportion of regular correspondence patterns and the highest proportion of regular words. The core-orientied trimming strategy outperforms the baseline without trimming in some cases, but not consistently, often only leading to minimal improvements over the baseline. Random tests confirm this trend for both trimming strategies.

The reduction of alignment sites generally leads to a reduced number of correspondence patterns in-ferred from the individual datasets, no matter which trimming procedure is applied. This holds in all settings for both irregular and regular correspondence patterns (see Appendix A for details). Gap-oriented trimming removes more patterns than core-oriented trimming, which is also expected, given that in the latter setting we preserve some sites in the core that would otherwise have been trimmed. Figure 4 visualizes the reduction of correspondence patterns and alignment sites for all ten datasets in our sample. This analysis allows us to make two general observations. First, frequently recurring correspondence patterns tend to grow with respect to the number of alignment sites in which they recur after trimming. We attribute this to the greedy nature of the correspondence pattern inference procedure. Second, the long tail of correspondence patterns with very few alignment sites is substantially shortened in almost all languages. This provides yet another perspective on the necessity of trimming in linguistics. Many of the patterns with a low amount of alignment sites do indeed seem to contain erroneous alignment judgements, and trimming them successfully improves the distribution of sites across the patterns. The two datasets where the tail does not seem substantially shortened, CROSSANDEAN and ZHIVLOVOBUGRIAN, are also the ones with the lowest gain in the proportion of regular correspondence patterns. While there are still small improvements, it does seem that in those cases the gap-oriented trimming does not seem as effective as for other datasets.

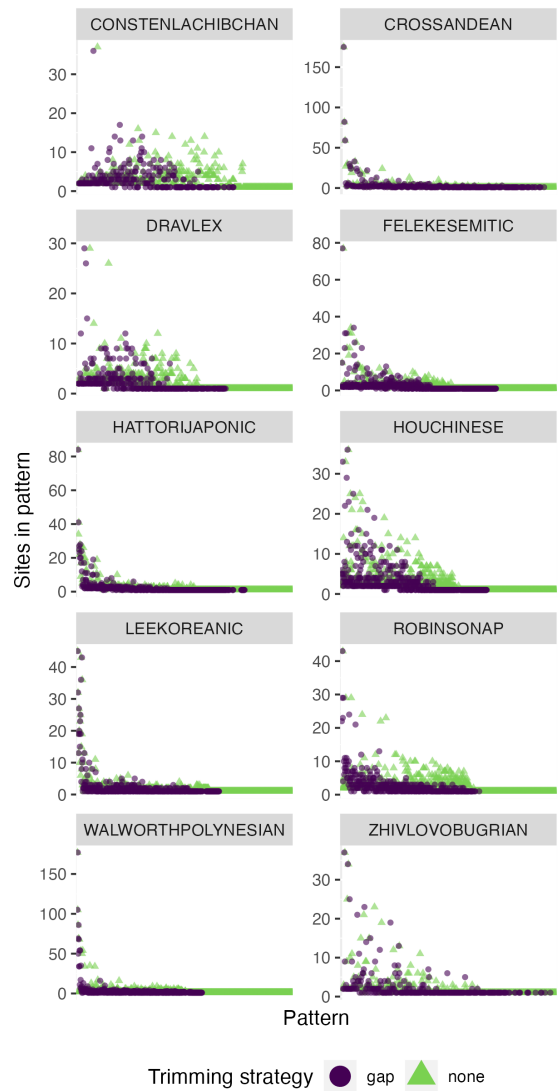

Figure 4: Distribution of alignment sites per pattern with gap-oriented trimming and without. Each point on the x-axis represents one correspondence pattern, its value on the y-axis reflects the number of alignment sites it contains. The patterns are sorted on the x-axis by their number of alignment sites. Gap-oriented trimming and the baseline are distinguished by shape and color.

|  | Original | | Core | | Gap | |
|---|---|---|---|---|---|---|
| Dataset | P | W | P | W | P | W |
| CONSTENLACHIBCHAN | 0.71 | 0.50 | 0.69/ 0.70 | 0.46/ 0.47 | **0.76**/ 0.70 | **0.51**/ 0.43 |
| CROSSANDEAN | 0.73 | 0.58 | 0.74/ 0.73 | 0.60/ 0.59 | **0.75**/ 0.73 | **0.64**/ 0.59 |
| DRAVLEX | 0.56 | 0.23 | 0.57/ 0.55 | 0.27/ 0.23 | **0.61**/ 0.55 | **0.31**/ 0.24 |
| FELEKESEMITIC | 0.55 | 0.22 | 0.58/ 0.56 | 0.25/ 0.24 | **0.62**/ 0.56 | **0.29**/ 0.25 |
| HATTORIJAPONIC | 0.58 | 0.33 | 0.57/ 0.58 | 0.33/ 0.33 | **0.59**/ 0.58 | **0.38**/ 0.34 |
| HOUCHINESE | 0.65 | 0.40 | 0.65/ 0.65 | 0.42/ 0.40 | **0.69**/ 0.64 | **0.45**/ 0.35 |
| LEEKOREANIC | 0.44 | 0.21 | 0.47/ 0.45 | 0.20/ 0.21 | **0.52**/ 0.47 | **0.22**/ 0.20 |
| ROBINSONAP | 0.64 | 0.36 | 0.65/ 0.63 | 0.37/ 0.47 | **0.67**/ 0.63 | **0.41**/ 0.35 |
| WALWORTHPOLYNESIAN | 0.66 | 0.40 | 0.66/ 0.65 | 0.40/ 0.39 | **0.72**/ 0.66 | **0.48**/ 0.39 |
| ZHIVLOVOBUGRIAN | 0.57 | 0.24 | 0.58/ 0.57 | 0.26/ 0.25 | **0.61**/ 0.58 | **0.28**/ 0.26 |

Table 2: Proportion of regular correspondence patterns (P) and regular words (W) across all datasets after trimming. The numbers after the slashes provide the average from 100 iterations of the random model.

One likely explanation for this observation is the fact that both datasets, as well as HATTORIJAPONIC, include language varieties that are closely related to each other. ZHIVLOVOBUGRIAN includes data from one subgroup of the Uralic language family, while the Quechua languages from CROSSANDEAN are generally considered to be quite similar to each other and of shallow time-depth. In those cases, we expect many forms that are (nearly) identical to each other. This would directly result in correspondence patterns of high frequency, from which not too many sites are trimmed. Especially for CROSSANDEAN, this is reflected by the fact that it has the highest proportion of regular words across all the datasets, pointing to a very regular set of lexical items.

Table 3 shows the results of our error analysis, comparing in how many out of 100 trials for each trimming strategy the proportion of regular words was higher in the random trial than in the concrete trimming method. As we can see from the table, the random-deletion model often outperforms the core-oriented trimming strategy, while it performs consistently worse than the gap-oriented trimming strategy. This clearly shows that it is not enough to trim alignment sites at random in order to reduce the noise in the data. As can be expected due to traditional theories on the regularity of sound change, specific sites, which reflect irregular correspondence patterns, must be targeted. For some datasets, the random model does surprisingly well in the core-oriented setting, and in some cases, it is even consistently better than the targeted core-strategy. This can be explained by the fact that the random trimming might also trim sites within the core – sites that apparently are very irregular in some languages – and hence improve the model in comparison to a trimming-model where a certain core is always preserved. Given that the model performs worse than the gap-oriented trimming in all languages, it seems recommendable to trim all sites above the gap-threshold, regardless of their position in the alignment. The successful trimming of sites that include a majority of gaps shows that those sites contain many irregular correspondences, and removing them improves our measures of regularity. We are now able to explain more words in the dataset with a lower number of regular correspondence patterns.

| Dataset | Core | Gap |
|---|---|---|
| CONSTENLACHIBCHAN | 0.58 | 0.00 |
| CROSSANDEAN | 0.02 | 0.00 |
| DRAVLEX | 0.00 | 0.00 |
| FELEKESEMITIC | 0.17 | 0.01 |
| HATTORIJAPONIC | 0.40 | 0.00 |
| HOUCHINESE | 0.05 | 0.00 |
| LEEKOREANIC | 0.54 | 0.06 |
| ROBINSONAP | 0.34 | 0.00 |
| WALWORTHPOLYNESIAN | 0.11 | 0.00 |
| ZHIVLOVOBUGRIAN | 0.12 | 0.05 |

Table 3: Percentage of models with random deletion of alignment sites that achieved higher regularity than the respective trimming model.

| | | | | | | | | | |
|---|---|---|---|---|---|---|---|---|---|
| Boruca | - | - | b | ɾ | u | - | ŋ | - | - | - |
| Cabecar | - | - | b | - | u | - | ɹ | i | t | u |
| Chimila | - | - | b | - | u | h | ŋ | a | ʔ | - |
| Malayo | - | - | b | - | ɨ | - | n | - | - | - |
| Ngabere | ŋ | ɯ | b | ɾ | ɯ | - | - | - | - | - |
| Proto-Chibchan | | | ᵐb | | ũ | | ⁿd | | | |

Figure 5: Gap-oriented trimming for the cognate words of ASHES in Chibchan languages

| | | | |
|---|---|---|---|
| Boruca | d | i | ʔ |
| Bribri | d | i | ʔ |
| Buglere | tʃ | i | - |
| Cogui | n | i | - |
| Ngabere | ɲ | ɤ | - |
| Proto-Chibchan | ⁿd | i | ʔ |

Figure 6: Trimming for the cognate words of WATER in Chibchan

Further experimentation will have to be done with respect to different gap thresholds. Our initial threshold of 0.5 reflects the fact that we did not want to search for the threshold with the highest number of regularity, but rather to account heuristically for sites that include more gaps than reflexes of sound. Furthermore, the optimal threshold might well be different for each language family, given that correspondence patterns can differ greatly across languages. For example, patterns of change in which sounds are lost in certain positions might be very frequent for one language family, but not in another, leading to a different role of gaps in the correspondence patterns.

### 4.2 Success and Failure of Trimming

Our implementation is fully compatible with computer-assisted workflows (List, 2017b). We output all data in a way that experts can check them, and make both the trimmed sites as well as the resulting (ir)regular correspondence patterns explicit. This makes it possible to use the output of our method in various tasks in historical linguistics. Figure 5 provides one example from the CONSTENLACHIBCHAN dataset of the output that our trimming provides. The figure presents a subset of cognate words for the concept ASHES, including all gaps in the original alignment from the selected languages. All alignment sites which featured mostly gaps were successfully trimmed from the alignment and are displayed as greyed out in the example. Three alignment sites remain, which pattern well with the reconstruction of ASHES in Proto-Chibchan as provided by Pache (2018, 41). If the core-oriented trimming were performed instead, five instead of three alignment sites would have remained in the final alignment, as the two sites represented by the fourth and sixth column are within the preserved core. This case illustrates the advantage of the gap-oriented trimming strategy, as all spurious alignment sites are trimmed from the data, regardless of their position.

The closer inspection of individual trimmed alignments shows that our methods still have a lot of room for improvement. One major problem lies in the nature of the gap-oriented trimming. As we remove all sites which include mostly gaps, we might lose relevant correspondence patterns in which the gaps do not constitute an erroneous alignment, but rather an actual case of gaps in the pattern. It is a very reasonable assumption that there are language families in which merger with zero occurred for some correspondence pattern in the majority of languages. One such example can be found in Figure 6, where the trimmed alignments for the concept WATER in several Chibchan languages can be found. Again, we add to the data from the CONSTENLACHIBCHAN-dataset the reconstruction as provided by Pache (2018, 235). As we can see, the alignment site which includes the reflexes the glottal stop as reconstructed for Proto-Chibchan contains gaps in most languages. With the current methodology which focuses exclusively on gaps, this pattern will be trimmed from the alignment, despite reflecting relevant information. This is paralleled by discussions in biology, where gaps might contain phylogenetically relevant information (Tan et al., 2015). This opens up the question whether we will be able to feed such information into the trimming algorithm, and preserve certain patterns that we know of that would otherwise be trimmed.

What remains to be done in future studies is to manually evaluate trimmed correspondence patterns. This is a general task for historical language comparison, as linguists often base their reconstruction judgements on impressionistic statements of regularity or only report the most frequent correspondence patterns.

## 5 Conclusion

We introduce the concept of trimming multiple sequence alignments, originally developed for applications in evolutionary biology, to the field of historical linguistics. Trimming as such is already practiced implicitly in the comparative method, but as of yet, there are no computational implementations for the procedure. Our trimming algorithms provide considerable improvements compared to state-of-the-art alignment methods. By trimming the alignment sites down to a subsequence without gaps, we achieve a higher number of regular correspondence patterns and cognate sets than without trimming. Even though our technique is merely a very preliminary approximation to the classical workflow of the comparative method, the average regularity of correspondence patterns across data sets is improved in all settings analyzed. Our study thus shows that automated trimming is both achievable and worthwhile in computational historical linguistics.

The main target of our trimming-strategies were alignment sites that included more gaps than defined in a certain threshold. Our model comparison shows that the best results are achieved when all such sites are trimmed, rather than only those at the periphery of stable alignment sites. Similar to biology, we find that alignment sites with many gaps contain divergent information, and trimming them improves the accuracy of our methods. It is also not sufficient to trim sites at random, since in that case we lose correspondence patterns that explain the data well. The examples we provide show both the potential of trimming alignment sites and their methodological limitations. The success of our strategy varies considerably between the datasets. A closer analysis of those cases where improvements are considerably small could provide valuable information for improved trimming strategies to be implemented in the future.

## Limitations

In addition to the already discussed problems related to the exclusive focus on gaps, we have only tested the trimming with respect to a generalized function of regularity in each dataset. It is not yet clear whether this actually improves the computational success of secondary tasks like reconstructions or new methods of cognate detection.

## Ethics Statement

Our data are taken from publicly available sources. For this reason, we do not expect that there are ethical issues or conflicts of interest in our work.

## Supplementary Material

The supplementary material accompanying this study contains the data and code needed to replicate the results reported here, along with detailed information on installing and using the software. It is curated on GitHub (https://github.com/pano-tacanan-history/trimming-paper, Version 1.1) and has been archived with Zenodo (https://doi.org/10.5281/zenodo.7780719).

## Acknowledgements

This research was supported by the Max Planck Society Research Grant *CALC³* (FB, JML, https://digling.org/calc/) and the ERC Consolidator Grant *ProduSemy* (JML, Grant No. 101044282, see https://doi.org/10.3030/101044282). Views and opinions expressed are however those of the authors only and do not necessarily reflect those of the European Union or the European Research Council Executive Agency (nor any other funding agencies involved). Neither the European Union nor the granting authority can be held responsible for them. We thank Nathan W. Hill and Thiago C. Chacon and the anonymous reviewers for helpful comments. We are grateful to all people who share their data openly.

## A  Table of Results for Individual Datasets

| Analysis | Frequ. Pat. | Rare Pat. | All Pat. | Reg. Words | Irr. Words | All Words |
|---|---:|---:|---:|---:|---:|---:|
| constenlachibchan | 884 | 355 | 1239 | 607 | 609 | 1216 |
| constenlachibchan/gap | 593 | 188 | 781 | 622 | 594 | 1216 |
| constenlachibchan/gap/r | 549 | 232 | 781 | 517 | 699 | 1216 |
| constenlachibchan/core | 680 | 304 | 984 | 563 | 653 | 1216 |
| constenlachibchan/core/r | 693 | 291 | 984 | 572 | 644 | 1216 |
| crossandean | 781 | 296 | 1077 | 1624 | 1165 | 2789 |
| crossandean/gap | 724 | 243 | 967 | 1777 | 1012 | 2789 |
| crossandean/gap/r | 708 | 259 | 967 | 1660 | 1129 | 2789 |
| crossandean/core | 769 | 276 | 1045 | 1667 | 1122 | 2789 |
| crossandean/core/r | 760 | 285 | 1045 | 1634 | 1155 | 2789 |
| dravlex | 665 | 515 | 1180 | 312 | 1029 | 1341 |
| dravlex/gap | 494 | 311 | 805 | 415 | 926 | 1341 |
| dravlex/gap/r | 439 | 366 | 805 | 317 | 1024 | 1341 |
| dravlex/core | 591 | 442 | 1033 | 359 | 982 | 1341 |
| dravlex/core/r | 566 | 466 | 1033 | 306 | 1035 | 1341 |
| felekesemitic | 928 | 755 | 1683 | 579 | 2043 | 2622 |
| felekesemitic/gap | 824 | 504 | 1328 | 773 | 1849 | 2622 |
| felekesemitic/gap/r | 743 | 585 | 1328 | 643 | 1979 | 2622 |
| felekesemitic/core | 860 | 632 | 1492 | 654 | 1968 | 2622 |
| felekesemitic/core/r | 838 | 654 | 1492 | 632 | 1990 | 2622 |
| hattorijaponic | 812 | 580 | 1392 | 562 | 1148 | 1710 |
| hattorijaponic/gap | 620 | 424 | 1044 | 644 | 1066 | 1710 |
| hattorijaponic/gap/r | 600 | 444 | 1044 | 587 | 1123 | 1710 |
| hattorijaponic/core | 707 | 534 | 1241 | 569 | 1141 | 1710 |
| hattorijaponic/core/r | 721 | 520 | 1241 | 568 | 1142 | 1710 |
| houchinese | 1329 | 726 | 2055 | 723 | 1093 | 1816 |
| houchinese/gap | 1020 | 453 | 1473 | 819 | 997 | 1816 |
| houchinese/gap/r | 940 | 533 | 1473 | 640 | 1176 | 1816 |
| houchinese/core | 1212 | 646 | 1858 | 756 | 1060 | 1816 |
| houchinese/core/r | 1201 | 657 | 1858 | 723 | 1093 | 1816 |
| leekoreanic | 603 | 764 | 1367 | 441 | 1690 | 2131 |
| leekoreanic/gap | 524 | 480 | 1004 | 464 | 1667 | 2131 |
| leekoreanic/gap/r | 467 | 537 | 1004 | 433 | 1698 | 2131 |
| leekoreanic/core | 543 | 623 | 1166 | 434 | 1697 | 2131 |
| leekoreanic/core/r | 521 | 645 | 1166 | 440 | 1691 | 2131 |
| robinsonap | 1094 | 616 | 1710 | 518 | 906 | 1424 |
| robinsonap/gap | 742 | 358 | 1100 | 584 | 840 | 1424 |
| robinsonap/gap/r | 693 | 407 | 1100 | 498 | 926 | 1424 |
| robinsonap/core | 877 | 479 | 1356 | 532 | 892 | 1424 |
| robinsonap/core/r | 861 | 495 | 1356 | 523 | 901 | 1424 |
| walworthpolynesian | 1568 | 820 | 2388 | 1472 | 2165 | 3637 |
| walworthpolynesian/gap | 1187 | 470 | 1657 | 1746 | 1891 | 3637 |
| walworthpolynesian/gap/r | 1094 | 563 | 1657 | 1414 | 2223 | 3637 |
| walworthpolynesian/core | 1377 | 708 | 2085 | 1452 | 2185 | 3637 |
| walworthpolynesian/core/r | 1357 | 728 | 2085 | 1415 | 2222 | 3637 |
| zhivlovobugrian | 469 | 355 | 824 | 482 | 1492 | 1974 |
| zhivlovobugrian/gap | 414 | 265 | 679 | 546 | 1428 | 1974 |
| zhivlovobugrian/gap/r | 393 | 286 | 679 | 506 | 1468 | 1974 |
| zhivlovobugrian/core | 420 | 307 | 727 | 505 | 1469 | 1974 |
| zhivlovobugrian/core/r | 413 | 314 | 727 | 494 | 1480 | 1974 |

Table 4: Full results with information on all patterns and words